\documentclass[runningheads]{llncs}
\usepackage{graphicx,hyperref}
\usepackage{microtype}
\usepackage{booktabs}
\usepackage{multirow}
\usepackage{svg}

\usepackage{amsmath,amsfonts,bm}

\def\eqref#1{equation~\ref{#1}}

\def\1{\bm{1}}

\def\ve{{\bm{e}}}

\def\vh{{\bm{h}}}

\def\vp{{\bm{p}}}

\def\vt{{\bm{t}}}

\def\vw{{\bm{w}}}

\def\mH{{\bm{H}}}

\def\mP{{\bm{P}}}

\def\mT{{\bm{T}}}

\def\mW{{\bm{W}}}

\DeclareMathAlphabet{\mathsfit}{\encodingdefault}{\sfdefault}{m}{sl}
\SetMathAlphabet{\mathsfit}{bold}{\encodingdefault}{\sfdefault}{bx}{n}

\def\gA{{\mathcal{A}}}

\def\gE{{\mathcal{E}}}

\def\gL{{\mathcal{L}}}

\def\gN{{\mathcal{N}}}

\def\gS{{\mathcal{S}}}

\def\gV{{\mathcal{V}}}
\def\gW{{\mathcal{W}}}

\DeclareMathOperator*{\enc}{Trans-Enc}
\DeclareMathOperator*{\dec}{Trans-Dec}
\DeclareMathOperator*{\cls}{[CLS]}
\DeclareMathOperator*{\sep}{[SEP]}
\DeclareMathOperator*{\mlp}{MLP}
\DeclareMathOperator*{\prompt}{Summarization~by}

\DeclareMathOperator*{\relu}{ReLU}

\begin{document}

\title{Topic-Selective Graph Network for Topic-Focused Summarization}

\author{Zesheng Shi$^1$, Yucheng Zhou$^2$}

\authorrunning{Z.Shi and Y.Zhou}

\institute{$^1$Nankai University \ \  $^2$University of Technology Sydney \\
\email{2120210083@mail.nankai.edu.cn}, \email{yucheng.zhou-1@student.uts.edu.au}}

\maketitle              

\begin{abstract}
Due to the success of the pre-trained language model (PLM), existing PLM-based summarization models show their powerful generative capability. 
However, these models are trained on general-purpose summarization datasets, leading to generated summaries failing to satisfy the needs of different readers.
To generate summaries with topics, many efforts have been made on topic-focused summarization.
However, these works generate a summary only guided by a prompt comprising topic words.
Despite their success, these methods still ignore the disturbance of sentences with non-relevant topics and only conduct cross-interaction between tokens by attention module.
To address this issue, we propose a topic-arc recognition objective and topic-selective graph network.
First, the topic-arc recognition objective is used to model training, which endows the capability to discriminate topics for the model.
Moreover, the topic-selective graph network can conduct topic-guided cross-interaction on sentences based on the results of topic-arc recognition.
In the experiments, we conduct extensive evaluations on NEWTS and COVIDET datasets.
Results show that our methods achieve state-of-the-art performance.
\end{abstract}

\section{Introduction}
Text summarization aims to compress a long article into a short and clear summary, which is a fundamental task in many NLP applications.
With the success of sequence-to-sequence (seq2seq) language models, it is widely integrated into many real-world applications, e.g., document snippets generation in search engines \cite{Turpin07Fast}, automatic news summaries \cite{Ahuja22ASPECTNEWS} and legal document summarization \cite{Polsley16CaseSummarizer}. 
In recent years, text summarization has been an essential area in academia and industry.

With advanced deep learning, summarization models are generally designed based on the seq2seq framework \cite{cho2014learning}.
Recently, many pre-trained language models (PLM)  are proposed by pre-training a Transformer \cite{vaswani2017attention} in a large-scale unlabeled corpus in a self-supervised manner. 
Since the PLM encapsulates large-scale language prior knowledge, it shows an excellent generative capability on many natural language generation tasks, e.g., caption generation \cite{Zhou22Sketch}, machine translation \cite{Ren20SemFace} and event generation \cite{Zhou22ClarET}. 
Therefore, PLM-based models also have become a mainstream paradigm for text summarization.  
However, since the training summarization model by only finetuning is still somewhat insufficient, researchers propose many methods to improve the PLM-based summarization model, e.g., contrastive learning \cite{su2022contrastive} and information retrieval \cite{Bouras08Improving}.  

Although it is very successful in text summarization based on PLM, these methods suffer from a non-focused problem.
Since most existing summarization models are trained on general-purpose datasets, they focus on generating general-purpose summaries.
A general-purpose summary fails to satisfy the needs of different readers and reflects the full range of content of the article.
Recently, many text generation methods focus on controllable generation processes guided by sentiment polarity \cite{titov-mcdonald-2008-joint} and specific topic distributions \cite{liu2019topic}.
However, these methods lack effective evaluation due to no topic-focused summarization dataset existing.
Therefore,  {\it Bahrainian et al.}\cite{bahrainian-etal-2022-newts} introduce a NEWs Topic-focused Summarization (NEWTS) corpus to close this gap and propose prompt-based methods to improve PLM-based summarization methods.

\begin{figure}[t]
\centering
\includegraphics[width=\textwidth]{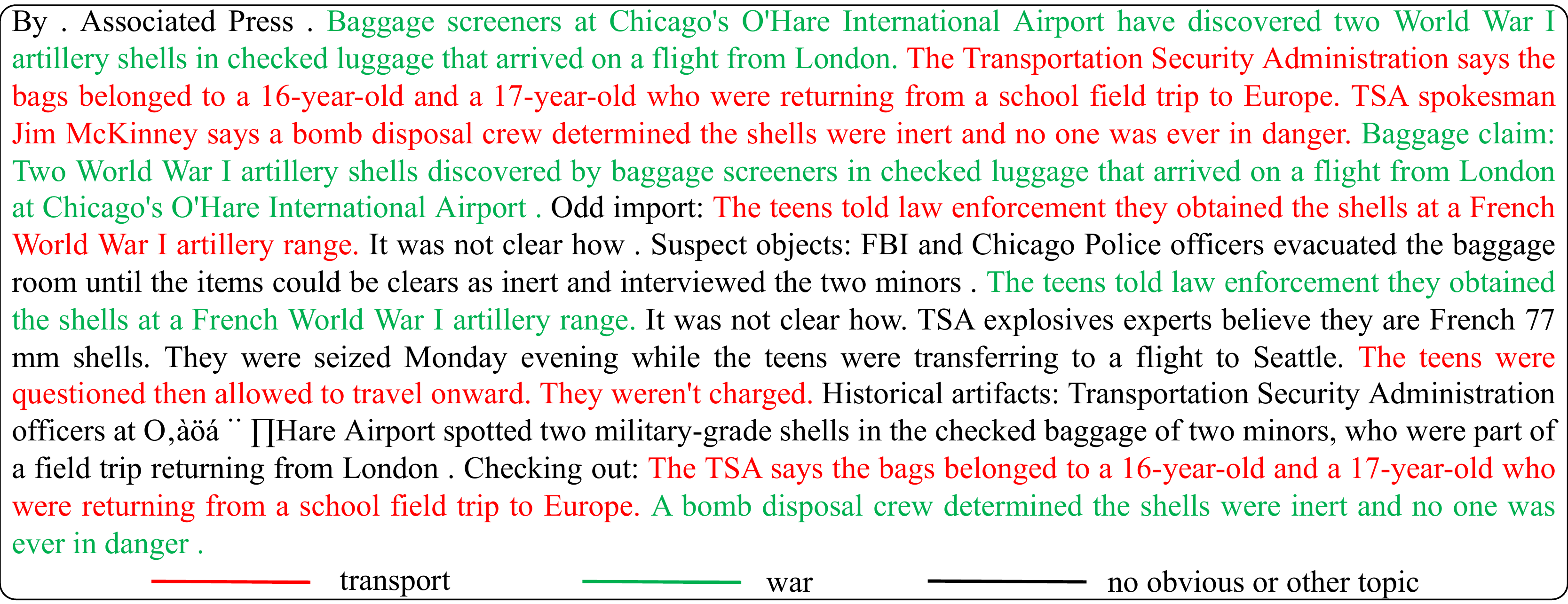}
\caption{Sentences with multiple topics in an article.}
\label{fig:intro}
\end{figure}

Despite the success of these works, they neglect topic-guided cross-interaction between sentences. 
As shown in Figure \ref{fig:intro}, there are multiple sentences with different topics in an article. 
To generate a summary more relevant to a given topic, a topic-focused summarization model is required to distinguish topics in sentences and conducts cross-interaction between sentences with the given topic.
In contrast, existing methods \cite{shin2020autoprompt} generate a summary only guided by a prompt comprising topic words, which ignores the disturbance of sentences with non-relevant topics and only conducts cross-interaction between tokens by attention module.

Due to sentences with multiple topics in an article, we propose a topic-arc recognition objective to distinguish the topic of each sentence in the article. 
Since the summary has a definite topic, we leverage summaries and their topics on an article to train the model that can predict the topic of sentences in the article.
Moreover, summaries selected from an article have a similar context, which pushes the model to focus on the topic instead of bias in content.
In addition, we propose a topic-selective graph network, which can conduct topic-guided cross-interaction on sentences in the article.
Specifically, we first construct a graph based on sentence and topic candidates sampled from prediction results of topic-arc recognition on the article. 
Then, the graph nodes are updated by relational graph convolution layers.
Lastly, the updated node representations are delivered to the decoder for summary generation.

In the experiments, we conduct extensive evaluations on two datasets, i.e., NEWTS \cite{bahrainian-etal-2022-newts} and COVIDET \cite{ZhanETAL22CovidET}. Experimental results show that our method outperforms other strong competitors and achieves state-of-the-art performance on NEWTS and COVIDET. Moreover, we further analyze the effectiveness of our method by providing additional quantitative and qualitative results.

\section{Related Work}
\subsection{PLM-based Summarization}
With the rise of seq2seq models, researchers are increasingly interested in text summarization. 
The original topic-based model was the TOPIARY model proposed by {\it Li et al.}\cite{li2004topiary} in 2004, which combined language-driven compression techniques and unsupervised topic detection methods. 
With the advance of the pre-training technique \cite{radford2018improving}, the development of PTMs in text summarization is also booming \cite{li2021pretrained}. 
There are many sequence-to-sequence pre-trained language models that show their powerful capability for summarization, e.g., BART \cite{lewis2019bart}, T5 \cite{raffel2020exploring} and Prophetnet \cite{qi2020prophetnet}. 
In addition, to be better compatible with discriminative and generation tasks, {\it Dong et al.} \cite{dong2019unified} propose UniLM that can be used in natural language understanding and generation tasks. The model also shows excellent summarization capability. 

\subsection{Topic-Guided Summarization}
With the advance of text summarization, increasing researchers are interested in generating topic-specific summaries.
Initially, the LDA model is used to guide the topic of summary \cite{li2004topiary}. 
For example, {\it Xing et al.}\cite{xing2017topic} propose a topic aware seq2seq model named Twitter LDA for a response, which introduces topic information through a joint attention mechanism and a bias generation probability. 
Another work, CATS \cite{10.1145/3464299}, is a neural sequence-to-sequence model based on an attentional encoder-decoder architecture, which introduces a new attention mechanism named topic attention controlled by an unsupervised topic model. 
In PTM-based models, the Plug and Play Language Model (PPLM) \cite{dathathri2019plug} is based on GPT-2 \cite{radford2019language}. 
In addition, the BART-FT-JOINT proposed in \cite{ZhanETAL22CovidET} can simultaneously use a sentiment inducer for sentiment-specific summary generation.

\subsection{Graph Neural Network}
Graph neural network \cite{scarselli2004graphical} has been valued in the field of deep learning for its excellent processing ability on unstructured data and node-centric information aggregation mode. 
With the advance of graph neural networks, there are many graph networks with special structures, e.g., GCN \cite{schlichtkrull2018modeling}, GAT \cite{velickovic2017graph}, HAN \cite{yang2016hierarchical} and r-GCN \cite{schlichtkrull2018modeling}.
Moreover, GNN is often used for downstream tasks such as text classification, information extraction, and text generation.
In text summarization, {\it Wang et al.}\cite{wang2020heterogeneous} propose a heterogeneous graph-based neural network for extracting summaries, which contains semantic nodes of different granularity levels except sentences. 
These extra nodes act as ``intermediaries" between sentences and enrich cross-sentence relations. 
The introduction of document nodes allows the graph structure to be flexibly extended from a single document setup to multiple documents. Another work \cite{jing2021multiplex} proposes a multiplex graph summary (Multi-GraS) model based on multiplex graph convolutional networks that can be used to extract text summaries. This model not only considers Various types of inter-sentential relations (such as semantic similarity and natural connection), and intra-sentential relations (such as semantic and syntactic relations between words) are also modeled.

\section{Method}
This section starts with a base topic-focused summarization model, followed by our proposed methods, i.e., topic-arc recognition and topic-selective graph network. Lastly, we elaborate on the details of our model training.

\subsection{Base Topic-Focused Summarization Model}
Topic-focused summarization aims to generate a topic-relevant summary based on a long article.
Since pre-trained language models (PLMs) based on Transformer show powerful capability for text generation, a recent trend is to finetune a PLM as a summarization model.
In this work, we first introduce a base topic-focused summarization model based on PLM.
Specifically, given an article $\gA$ and topic words $\gW$ corresponding to topic $t$, we first use topic words as a prefix prompt for the article and pass them into the Transformer encoder, i.e.,
\begin{align}
    \mH = \enc(\cls \prompt \gW : \gA \sep), \label{equ:enc}
\end{align}
where $\mH$ denotes token representations generated by the Transformer encoder, and $\mH \in \{\vh_0, \vh_1, \cdots, \vh_l\}$. $l$ is the length of the input.

Next, we deliver the token representations $\mH$ and the gold summary $\gS$ into the Transformer decoder, i.e.,
\begin{align}
    \mP = \dec(\mH, \gS), \label{equ:dec}
\end{align}
where $\mP = \{\vp_0, \vp_1, \cdots, \vp_n\}$ and $\vp_i$ is a probability distribution over vocabulary $\gV$. $n$ is the length of the summary.

Lastly, we train the pre-trained Transformer by maximum likelihood estimation, and its loss function is defined as:
\begin{align}
    \gL^{(ce)} = - \dfrac{1}{|n|}\sum_{n}\log\vp_i{[y=s_i]}
\end{align}
where $s_i$ is $i$-th word in the ground truth summary $\gS$.

\subsection{Topic-Arc Recognition}
\begin{figure}[t]
\centering
\includegraphics[width=0.9\textwidth]{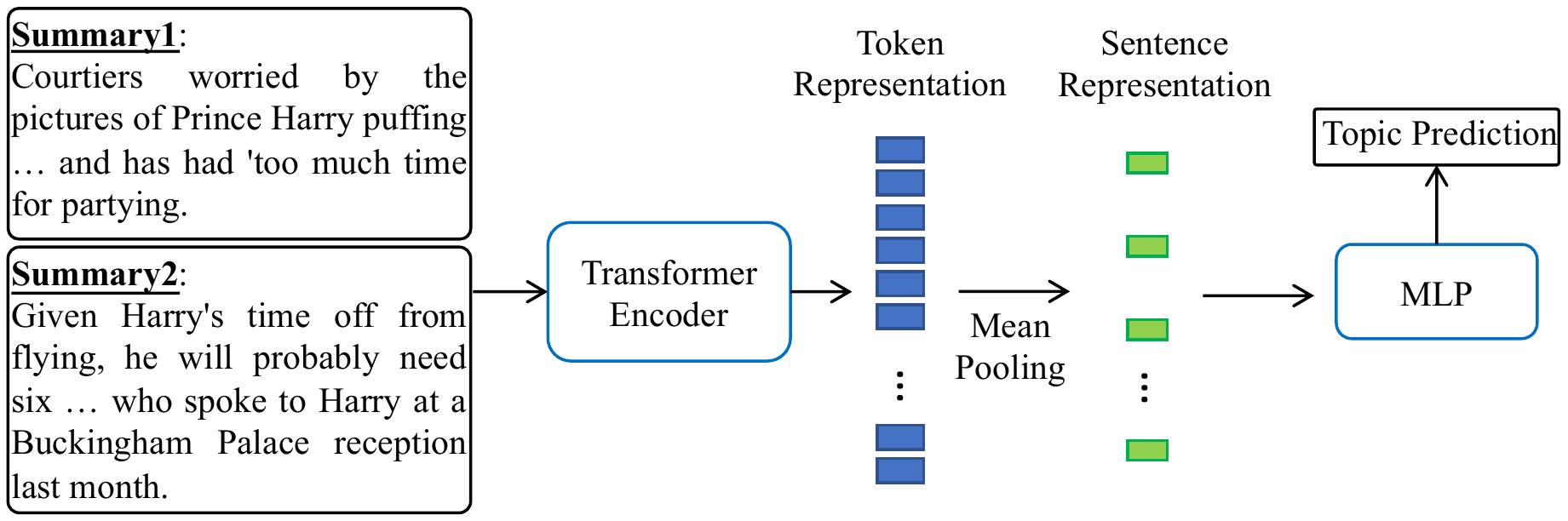}
\caption{Overview of topic-arc recognition.}
\label{fig:tar}
\end{figure}
Due to multiple sentences with different topics in an article, distinguishing topics of sentences in an article is essential to topic-focused summarization. 
To recognize the sentence's topic in articles, we propose a topic-arc recognition (TAR) objective, as shown in Figure \ref{fig:tar}.
In this objective, we concatenate two summaries ($\gS_0$ and $\gS_1$) with different topics ($t_1$ and $t_2$) from an article and pass them into the Transformer encoder, i.e.,
\begin{align}
    \mH = \enc(\gS_0, \gS_1).
\end{align}
Next, we conduct mean pooling to token representations $\mH$ to obtain sentence representations $\hat \mH = \{\hat \vh_1, \hat \vh_2, \cdots, \hat \vh_m\}, m = m_1 + m_2$, and $m_1$ and $m_2$ denote number of sentences in two summaries, respectively. 
Then, we deliver sentence representations $\hat \mH$ to a multilayer perceptron (MLP) to predict the topic of each sentence, i.e.,
\begin{align}
    \vp^t_i = \mlp(\hat \vh_i), i \in \{1,2, \cdots, m\}, \label{equ:mlp}
\end{align}
where $\vp^t_i$ denotes the topic probability distribution of $i$-th sentence on all topic categories.
Lastly, we optimize the Transformer encoder and MLP by a cross-entropy loss, i.e.,
\begin{align}
    \gL^{(tar)} = - \dfrac{1}{|m|}\sum_{m}\log\vp^t_i{[y=t_i]}, \label{equ:loss_tar}
\end{align}
where $t_i$ is the ground truth topic category of $i$-th sentence.

\subsection{Summarization with Topic-Selective Graph Network}
\begin{figure}[t]
\centering
\includegraphics[width=\textwidth]{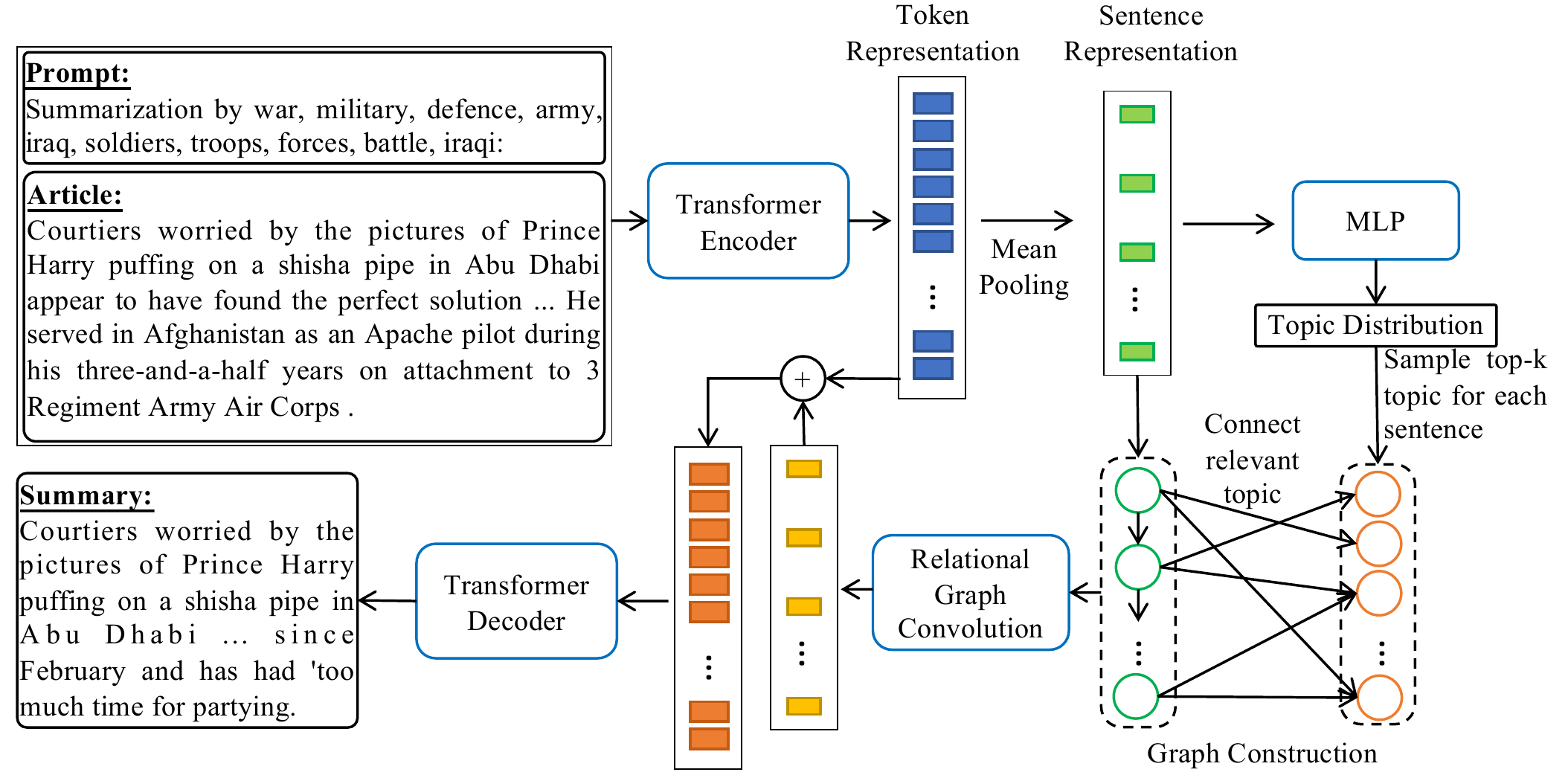}
\caption{Overview of summarization with topic-selective graph network.}
\label{fig:tsgn}
\end{figure}
As mentioned in Figure \ref{fig:intro}, sentences with the same topic in an article are usually not connected together.
To integrate semantic information on the same topic, we propose a topic-selective graph network (TSGN) to conduct sentence-level cross-interaction through topic nodes as bridge.
As shown in Figure \ref{fig:tsgn}, we first extract token representations $\mH$ in an article via Equ.\ref{equ:enc}.
Next, we conduct a mean pooling operation to token representations $\mH$ to obtain sentence representations $\bar \mH = \{\bar \vh_1, \bar \vh_2, \cdots, \bar \vh_m\}$.
Then, we deliver sentence representations $\bar \mH$ to the MLP in Equ.\ref{equ:mlp} to predict the topic probability distribution $\bar \vp^t_i$ of each sentence.
According to the ranking of probability $\bar \vp^t_i$, we select top-$k$ topics for each sentence as its topic candidates.

To conduct topic-guided cross-interaction, we construct a semantic graph, and sentence and topic nodes consist of sentence representations $\bar \mH$ and topic representations $\mT = \{\vt_1, \vt_2, \cdots, \vt_v\}$ that are derived from an embedding layer; and $v$ is number of topic categories.
Due to heterogeneous nodes of sentence and topic, we connect these nodes through edges with different relations $\gE \in \{\ve_s, \ve_t\}$.
The semantic graph comprises two rules of connecting nodes: 
1) We take the edge $\ve_s$ to connect sentence nodes $\bar \vh_i$ in sentence order. 
2) Each sentence node $\bar \vh_i$ is only connected to its topic candidate nodes $\vt_u$ through edge $\ve_t$, which can alleviate the disturbance of sentences with non-relevant topics.
Moreover, we use relational graph convolution layer \cite{schlichtkrull2018modeling,Zhou23Multimodal} to update sentence representations, i.e.,
\begin{align}
\vw^{l+1}_i = \relu \Big(\sum_{e \in \gE} \sum_{j \in \gN_{e}(i)} \frac{1}{\left|\gN_{e}(i)\right|} \mW_{e} \cdot \vw_{j}^{l}\Big), \vw_{j}^{0} \in \{\bar \mH, \mT\} \label{equ:rgcn}
\end{align}
where $\gE$ denote a set of all edges types, and $\gN_{e}(i)$ is the neighborhood of node $i$ under relation $e$. $l$ is number of the relational graph convolution layer. Updated sentence representations are represented as $\widetilde \mH$.
Furthermore, we take $\widetilde \mH$ plus into the corresponding token representations and pass them to Equ. \ref{equ:dec} to predict probability distributions $\widetilde \mP = \{\widetilde \vp_0, \widetilde \vp_1, \cdots, \widetilde \vp_n\}$. $n$ is length of the summary.
Lastly, we train our model by cross-entropy loss, which is defined as:
\begin{align}
    \gL^{(tsgn)} = - \dfrac{1}{|n|}\sum_{n}\log\widetilde \vp_i{[y=s_i]}
\end{align}
where $s_i$ is $i$-th word in the ground truth summary $\gS$.

\subsection{Training}
During training, we train our model by jointly topic-arc recognition and topic-selective graph network and minimize their loss functions:
\begin{align}
L = \alpha  \times \gL^{(tar)} + \beta  \times \gL^{(tsgn)} \label{equ:loss}
\end{align}

\section{Experiments}
\subsection{Dataset and Evaluation Metrics}
In the experiments, we train and evaluate our approach on two datasets, i.e., NEWTS \cite{bahrainian-etal-2022-newts} and COVIDET \cite{ZhanETAL22CovidET}. 
The NEWTS dataset is based on the famous CNN/Dailymail dataset and annotated by online crowd-sourcing. 
Every source article is paired with two summaries focusing on different topics and provides topic words to denote the topic.
The dataset consists of 2,400 and 600 samples in training and test sets. The results of the test set are reported by ROUGE (i.e., R-1, R-2, R-L) \cite{lin-2004-rouge} and Topic Focus \cite{bahrainian-etal-2022-newts}.
The COVIDET dataset is sourced from 1,883 English Reddit posts about the COVID-19 pandemic. Each post is annotated with 7 fine-grained emotion labels; for each emotion, annotators provided a concise, abstractive summary describing the triggers of the emotion.
We follow the official data split with 2,234/1,526 samples in training/test sets. The evaluation metrics are ROUGE-L (R-L) and BERT Score \cite{zhang2019bertscore}.

\subsection{Experimental Setting}
The pre-trained transformer we used is BART-base. 
We use AdamW optimizer with learning rate of 5 $\times 10^{-5}$, and learning rate of the relational graph convolution layer is 1 $\times 10^{-4}$. 
The weight decay and the dropout rate are both 1.0. 
The maximum training epoch, sentence number and batch size are set to 3, 60 and 2.
The length of input and output are set to 1024 and 128. 
We use NLTK to separate sentences.
In Equ.\ref{equ:loss}, $\alpha$ and $\beta$ are set to 1.0 and 0.8, respectively. 
Experiments are conducted on an NVIDIA RTX3080 GPU, and training time is around 3 hours.

\subsection{Main Results}
\begin{table}[t]
\centering
\caption{Results on NEWTS test set.}
\label{tab:new}
\begin{tabular}{lrrrc}
\toprule
\textbf{Method}   & \textbf{R-1}      & \textbf{R-2}      & \textbf{R-L}      & \textbf{Topic Focus} \\ \midrule
BART \cite{bahrainian-etal-2022-newts}  & 16.48            & 0.75             & 11.71            & 0.0080                \\
BART+T-W \cite{bahrainian-etal-2022-newts}       & 31.14            & 10.46            & 19.94            & 0.1375               \\
BART+CNN-DM \cite{bahrainian-etal-2022-newts}     & 26.23            & 7.24             & 17.12            & 0.1338               \\
T5+T-W \cite{bahrainian-etal-2022-newts}          & 31.78            & 10.83            & 20.54            & 0.1386               \\
T5+CNN-DM \cite{bahrainian-etal-2022-newts}       & 27.87            & 8.55             & 18.41            & 0.1305               \\
ProphetNet+T-W \cite{bahrainian-etal-2022-newts}    & 31.91            & 10.8             & 20.66            & 0.1362               \\
ProphetNet+CNN-DM \cite{bahrainian-etal-2022-newts} & 28.71            & 8.53             & 18.69            & 0.1295               \\
PPLM \cite{bahrainian-etal-2022-newts}              & 29.63            & 9.08             & 18.76            & 0.1482               \\
Ours              & \textbf{34.24} & \textbf{12.65} & \textbf{23.08} & \textbf{0.1512} \\ \bottomrule
\end{tabular}
\end{table}
\begin{table}[t]
\centering
\caption{Results on COVIDET test set. ``Ang", ``Dis", ``Fea", ``Joy", ``Sad", ``Tru" and ``Ant" denote anger, anticipation, joy, trust, fear, sadness and disgust, respectively.}
\label{tab:covid}
\begin{tabular}{lllllllll}
\toprule
\textbf{Method} & \textbf{Ang} & \textbf{Dis} & \textbf{Fea}  & \textbf{Joy}   & \textbf{Sad} & \textbf{Tru} & \textbf{Ant} & \textbf{AVG.} \\ \midrule
\textit{Metric: R-L}    &       &       &       &       &                &       &                &       \\ \midrule
BART \cite{ZhanETAL22CovidET}            & 0.161 & 0.138 & 0.164 & 0.149 & 0.157          & 0.158 & 0.164          & 0.156 \\
PEGASUS-FT \cite{ZhanETAL22CovidET}     & 0.185 & 0.155 & 0.199 & 0.158 & 0.173          & 0.164 & 0.193          & 0.175 \\
BART-FT \cite{ZhanETAL22CovidET}        & 0.190  & 0.159 & 0.206 & 0.165 & \textbf{0.177} & 0.162 & \textbf{0.198} & 0.180  \\
BART-FT-JOINT \cite{ZhanETAL22CovidET}  & 0.190  & 0.158 & 0.203 & 0.163 & 0.175          & 0.165 & 0.196          & 0.179 \\
Ours            & \textbf{0.202} & \textbf{0.177}   & \textbf{0.223} & \textbf{0.208} & 0.166            & \textbf{0.201} & 0.186                 & \textbf{0.195}   \\ \midrule
\textit{Metric: BERT Score} &       &       &       &       &                &       &                &       \\ \midrule
BART\cite{ZhanETAL22CovidET}            & 0.587 & 0.558 & 0.529 & 0.551 & 0.559          & 0.571 & 0.558          & 0.559 \\
PEGASUS-FT \cite{ZhanETAL22CovidET}     & 0.681 & 0.713 & 0.739 & 0.683 & 0.705          & 0.663 & 0.736          & 0.703 \\
BART-FT \cite{ZhanETAL22CovidET}        & 0.705 & 0.695 & 0.748 & 0.699 & 0.718          & 0.653 & 0.749          & 0.710  \\
BART-FT-JOINT \cite{ZhanETAL22CovidET}  & 0.701 & 0.706 & 0.729 & 0.694 & 0.713          & 0.659 & 0.746          & 0.707 \\
Ours            & \textbf{0.885} & \textbf{0.880}    & \textbf{0.889} & \textbf{0.888} & \textbf{0.879}   & \textbf{0.881} & \textbf{0.881}        & \textbf{0.883}   \\ \bottomrule
\end{tabular}
\end{table}
Comparison results of our methods and other strong competitors are shown in Table \ref{tab:new} and Table \ref{tab:covid}.
From the tables, we find two observations. 
1) Methods using topic words as a prompt outperform that of no prompt, which demonstrates that the topic words as prompt are significant in generating topic-focused summarization.
2) our approach is superior to others and achieves state-of-the-art, demonstrating the effectiveness of our method. 
Meanwhile, it shows that making a topic-guided sentence-level cross-interaction can improve topic-focused summarization by capturing the topic-relevant content in the article.

\subsection{Ablation Study}
\begin{table}[t]
\centering
\caption{Ablation study of our approach.}
\label{tab:abl}
\begin{tabular}{lcccc}
\toprule
\textbf{Method} & \textbf{R-1}   & \textbf{R-2}   & \textbf{R-L}   & \textbf{Topic Foucs} \\ \midrule
Ours           & \textbf{34.24} & \textbf{12.65} & \textbf{23.08} & \textbf{0.1512}      \\
Ours w/o TSGN          & 34.12          & 12.08          & 22.61          & 0.1407               \\
Ours w/o TAR    & 34.30           & 11.97          & 22.27          & 0.1453               \\ 
Ours w/o TSGN, TAR       & 31.14            & 10.46            & 19.94            & 0.1375               \\\bottomrule
\end{tabular}
\end{table}
As shown in Table \ref{tab:abl}, we conduct an ablation study on our method. 
First, we remove the topic-arc recognition (TAR) objective, and the results show performance drops. It demonstrates that the TAR objective can distinguish the topic category of the sentence, which improve the capability of topic awareness in the model.
Moreover, we investigate the effectiveness of TSGN by removing it from our method. Results show that the performance drops, which demonstrates that TSGN can improve topic-focused summarization through cross-sentence interaction.
Lastly, we remove TAR and TSGN to verify the effectiveness of our method. The results show a large drop, which further supports the effectiveness of our method.

\subsection{Impact of Topic Node}
\begin{figure}[t]
\centering
\includegraphics[width=0.75\textwidth]{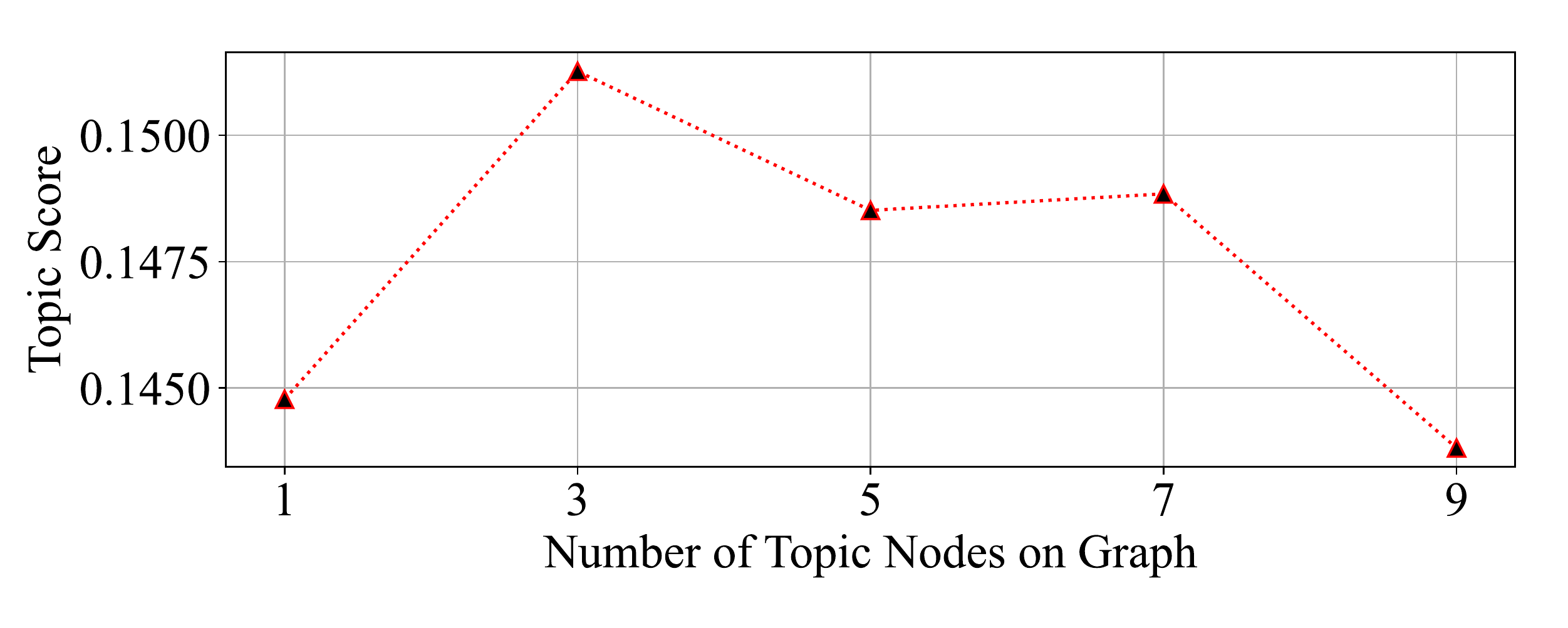}
\caption{Results on different number of topic nodes in topic-selective graph network.}
\label{fig:node}
\end{figure}
It can be seen from Figure \ref{fig:node} that the best performance is achieved when the number of topics selected in TSGN is 3. 
Moreover, the performance drops as the number of topic numbers increases or decreases. 
The reason is that the topic was sampled from the distribution with some noise.
Introducing too few topic nodes (e.g., 1 node) into TSGN leads to missing the correct topic.
In addition, to ensure that the correct topic node is introduced, the introduction of too many topic nodes (e.g., more than 3) leads to too much noise into the TSGN, which is also not conducive to the performance of the model.

\subsection{Case Study}
\begin{figure}[t]
\centering
\includegraphics[width=\textwidth]{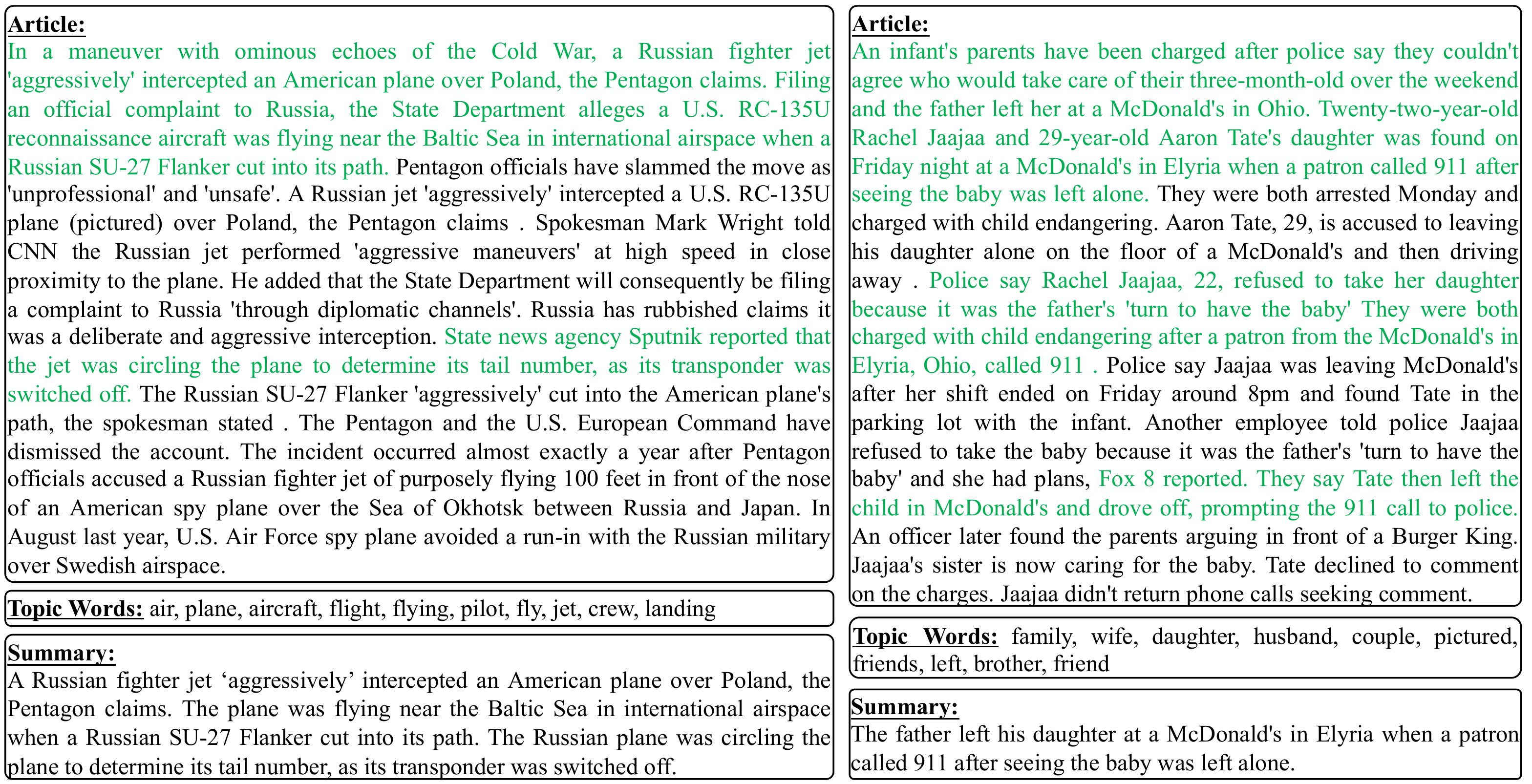}
\caption{Random sampling examples generated by our method.}
\label{fig:case}
\end{figure}
To conduct an extensive evaluation of our model, we take a qualitative comparison of our model. 
We randomly sample some generated summaries and their article and topic words, as are shown in Figure~\ref{fig:case}
In articles, the green area denotes that the sentence's topic is recognized as the given topic. 
Moreover, we can observe that the recognized sentences and given topics are relevant, which verifies the effectiveness of the TAR objective.
In addition, we can see that the generated summaries comprise the sentence content of the green area, which demonstrate the effectiveness of TSGN.

\subsection{Human Evaluation}
\begin{table}[t]
\centering
\caption{Human evaluation.}
\label{tab:hum}
\begin{tabular}{ccc}
\toprule
\multirow{2}{*}{\textbf{Type of evaluation}} & \multicolumn{2}{c}{\textbf{Win Percentage}} \\ \cline{2-3} 
                    & \textbf{BART+T-W} & \textbf{Ours} \\ \midrule
\bf Topic Relevance     & 0.27          & \textbf{0.73} \\
\bf Content Consistency & 0.36          & \textbf{0.64} \\
\bf Logic               & 0.41          & \textbf{0.59} \\ \bottomrule
\end{tabular}
\end{table}
To comprehensively evaluate our method, d, we conducted a human evaluation to compare our model and BART+T-W \cite{bahrainian-etal-2022-newts}.
We considered the topic relevance, content consistency and logic.
Therefore, there are three types of human evaluation. 
We randomly sampled 150 samples from the test set, and each sample includes an article, topic words and generated summary.
We displayed these samples to 3 recruited annotators. 
They need to distinguish which summaries are better quality based on the type of evaluation.
As shown in Table \ref{tab:hum}, results show that the performance of our model is significantly better than BART+T-W.

\section{Conclusion}
In this work, we dive into the limitations of previous topic-guided summarization methods, i.e., these methods still ignore the disturbance of sentences with non-relevant topics and only conduct cross-interaction between tokens by attention module.
 To address the limitations and improve the summarization model, we propose a topic-arc recognition objective and topic-selective graph network. The topic-arc recognition objective aims to discriminate topics of sentences. Moreover, the topic-selective graph network conducts topic-guided cross-interaction on sentences based on the results of topic-arc recognition. Experimental results show that our methods achieve state-of-the-art performance on NEWTS and COVIDET datasets.
 
\bibliographystyle{splncs04}
\bibliography{ref}
\end{document}